\documentclass[conference]{IEEEtran}
\IEEEoverridecommandlockouts

\usepackage{cite}
\usepackage{amsmath,amssymb,amsfonts}
\usepackage{algorithmic}
\usepackage{graphicx}
\usepackage{textcomp}
\usepackage{xcolor}
\usepackage[hidelinks]{hyperref}

\def\BibTeX{{\rm B\kern-.05em{\sc i\kern-.025em b}\kern-.08em
    T\kern-.1667em\lower.7ex\hbox{E}\kern-.125emX}}

\usepackage[breakable]{tcolorbox}
\usepackage{booktabs}
\usepackage{subcaption}
\usepackage{enumitem}
\usepackage{amsmath}
\usepackage{colortbl}
\usepackage{svg}
\usepackage{xspace}
\usepackage{url}
\usepackage{soul}
\usepackage{stfloats}
\graphicspath{ {./images/} }

\usepackage{orcidlink}

\newcommand{\second}[1]{\cellcolor{gray!20}{#1}}
\definecolor{lightgray}{gray}{0.92}

\newcommand{\tts}[0]{\textsc{Inter2US}\xspace}

\AtBeginEnvironment{tcolorbox}{\small}

\begin{document}

\title{Automated Alignment between Elicitation Interviews and Requirements}

\author{\IEEEauthorblockN{Francesco Dente}
\IEEEauthorblockA{\textit{Dept. of Data Science} \\
\textit{EURECOM}\\
Biot, France \\
francesco.dente@eurecom.fr\ }
\and
\IEEEauthorblockN{Fabiano Dalpiaz}
\IEEEauthorblockA{\textit{Dept. of Information and Computing Sciences} \\
\textit{Utrecht University}\\
Utrecht, The Netherlands \\
f.dalpiaz@uu.nl\ }
\and
\IEEEauthorblockN{Paolo Papotti}
\IEEEauthorblockA{\textit{Dept. of Data Science} \\
\textit{EURECOM}\\
Biot, France \\
paolo.papotti@eurecom.fr\ }
}

\maketitle

\begin{abstract}
Software requirements are derived from a variety of elicitation techniques, many of which have a conversational nature, like interviews. However, evaluating whether those derived requirements faithfully reflect the stakeholders’ needs remains a challenging manual task. 
In this paper, we formalize the task of aligning the transcript of an interview with a collection of requirements represented as user stories. We propose two heuristic metrics for alignment, called (i) \textit{requirements faithfulness}: the proportion of stories supported by the transcript, and (ii) \textit{interview coverage}: the proportion of transcript supported by at least one story. 
Then, we run experiments with large language models and embedding models that assess the ability of evaluating these metrics automatically. Experiments over four datasets show that an LLM-based solution achieves 0.86 macro-F1 on manually labeled chunk–story pairs. We also show how embedding models can be used as blockers to make the approach more scalable.
This work paves the way for more research on linking conversational artifacts with requirements. The formal framework and the automated matching techniques are basic components that can be used for emerging tasks such as tracing requirements to interviews and generating requirements from conversations. 

\end{abstract}

\begin{IEEEkeywords}
requirements interviews, user stories, alignment, traceability, large language models.
\end{IEEEkeywords}

\section{Introduction}
User stories are a frequently used artifact for expressing requirements in agile development~\cite{lucassen2016use}. These user-oriented requirements capture \textit{who} wants \textit{what} and \textit{why}~\cite{Cohn2004}. The most widespread notation for writing user stories is the Connextra template: ``\textit{As a [role], I want to [action], so that [benefit]}''.

User stories, like other types of requirements, are commonly distilled \textit{manually} from lengthy, conversational sources such as stakeholder interviews and facilitated workshops~\cite{wagner2019status}.

Recent research has leveraged Large language models (LLMs) to streamline this tedious human activity by generating candidate user stories from unstructured interview transcripts~\cite{ronanki2022chatgpt,quattrocchi2025can,yamani2025leveraging}, or even to explore LLM's ability to act as an elicitor who conducts interviews and derives requirements~\cite{korn2025llmrei}.

Both with human or automated generation, a research question (RQ) remains largely unaddressed: \emph{to what extent do the generated user stories faithfully reflect what stakeholders actually said?} 
Existing quality frameworks for user stories, like QUS~\cite{Lucassen2016QUS}, only assess how well a story is written, not whether it is \emph{grounded in the source}. Addressing this question is critical, because hidden and incomplete requirements are one of the most frequent causes for software project failure~\cite{fernandez2017naming}.

\begin{figure}[!h]
    \centering
    \includegraphics[width=0.85\columnwidth]{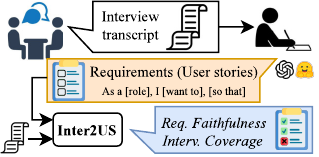}
    \caption{Overview of the workflow. Elicitation interviews produce a transcript. From this transcript, human analysts (or LLMs) craft user stories. Given a set of stories and an interview transcript, \tts computes two metrics: \emph{Requirements faithfulness} and \emph{Interview Coverage}.}
    \label{fig:over}
\end{figure}

We address this gap by introducing \tts, an NLP4RE task~\cite{zhao2021natural} that is concerned with measuring \textit{interview‑to‑story alignment}. The goal is to quantify alignment between an interview transcript and a set of user stories derived from it. 
We measure alignment via two interpretable, source‑faithful measures: (i) \emph{requirements faithfulness}, the proportion of stories supported by at least one segment of the transcript, and (ii) \emph{interview coverage}, the proportion of transcript segments that are covered by at least one story. As depicted in Fig.~\ref{fig:over}, given an interview's transcript and the set of corresponding stories, the goal is to output the two measures for 
an actionable quality assessment of the stories at hand. 

\begin{figure}[!h]
    \centering
    \includegraphics[width=\columnwidth]{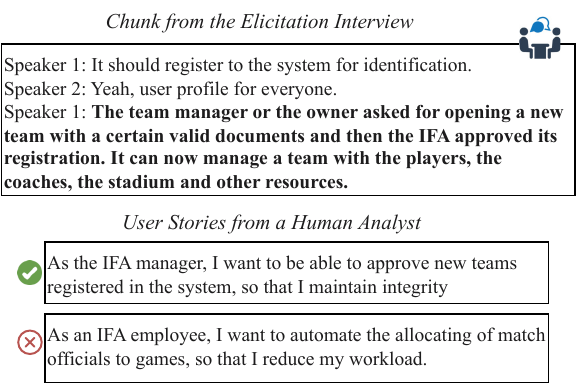}
\caption{Example of Interview-to-Story (\tts) alignment. A snippet from the elicitation interview (top) is \textit{matched} against two candidate user stories by a \emph{human analyst} (bottom). The former is a \textit{valid match}, the latter is \textit{not a match}. }
    \label{fig:ex}
\end{figure}

This paper makes four contributions that pave the way for research about the alignment of interview transcripts and generated requirements:
\begin{enumerate}
    \item We formalize \tts as a matching problem between conversation transcript \emph{chunks} and user stories, as depicted in Fig.~\ref{fig:ex}. Segmenting interviews into overlapping chunks enables local grounding and traceability: each positive match points to concrete evidence and can be surfaced as a justification during review~\cite{Spijkman2022Roots}. 
    \item  To make \tts practical, we introduce an \textit{embedding‑based blocking} scheme~\cite{blockingSurvey} that reduces the number of comparisons. Indeed, naively evaluating all story-chunk pairs scales quadratically and can become cost prohibitive for long interviews or large story sets.
    \item We conduct experiments on 17 software projects and show that an LLM-based matcher achieves an average macro-F1 of 0.86 on four manually annotated datasets for the matching task. 
    \item Using the top-performing matcher and the non-annotated datasets, we demonstrate that \emph{faithfulness} and \emph{coverage} can be used to compare generated user stories (both human- and LLM-written). 
\end{enumerate}

After discussing related work in Sect.~\ref{sec:relwork}, we introduce the \tts task in Sect.~\ref{sec:task}. We present our approach for addressing \tts in Sect.~\ref{sec:method}, describe the experimental setup in Sect.~\ref{sec:exp_setup}, and report on the results in Sect.~\ref{sec:results}. We conclude with a discussion of limitations and by sketching a research roadmap in Sect.~\ref{sec:discussion}.



\section{Related Work}
\label{sec:relwork}
We cast a common but under‑formalized software engineering activity into a precise NLP4RE task, involving methods from different research areas. 

\noindent \textbf{Requirements engineering (RE).}
User stories are central to agile RE~\cite{kassab_empirical_2014,lucassen2016use}, and their textual quality is typically assessed with frameworks such as QUS \cite{Lucassen2016QUS}.
While valuable, these criteria target how well a story is written (syntax, semantics, pragmatics) rather than whether the story is \emph{grounded} in the elicitation source.
Closer to our goal, Spijkman \textit{et al.}\ link user stories to elicitation conversations \cite{Spijkman2022Roots} and study how to summarize conversation transcripts to surface requirements-relevant turns \cite{spijkman2023summarization}.
\tts builds on this work by (i) formalizing alignment \emph{between} stories and interview chunks and (ii) turning grounding into explicit, coverage-style metrics rather than heuristics on isolated stories.

\noindent \textbf{Faithfulness and factuality.}
Existing work measures whether generated text is faithful to its source. In summarization,
hallucinations 
motivate source-grounded evaluation \cite{maynez-etal-2020-faithfulness,pagnoni-etal-2021-understanding,nenkova-passonneau-2004};
automatic metrics rely on NLI or QA signals to score consistency \cite{kryscinski-etal-2020-evaluating,laban-etal-2022-summac,wang-etal-2020-asking,scialom-etal-2021-questeval,fabbri-etal-2022-qafacteval,alignscore}.
\tts also employs the judge-against-the-source approach, but differs in unit and objective:
we align \emph{requirements} to \emph{conversational} evidence with many-to-many relations, and we report set-level coverage measures. 

Work on fact verification formalizes support/contradiction with retrieved evidence \cite{NakovCHAEBPSM21} and detects {previously fact-checked} claims with matching \cite{0002TNDP22,ShaarBMN20}.
Our setting differs as (i) claims (user stories) are longer, compositional artifacts rather than atomic statements, and
(ii) evidence lives in long, multi-speaker transcripts with local context and discourse.

\noindent \textbf{Retrieval, blocking, and reranking.}
Our \emph{embedding-based blocking} and \emph{pairwise matching} mirror standard information retrieval practice:
bi-encoders for fast similarity search \cite{reimers-gurevych-2019-sentence},
cross-encoder rerankers for precise scoring \cite{nogueira2019passage},
and 
techniques for reducing quadratic comparisons \cite{blockingSurvey}.
Unlike open-domain retrieval, we do not aim at answering a query but preserving alignment \emph{recall} at minimal token cost, so that a stronger judge can examine a pruned set of story–chunk pairs.

\noindent \textbf{LLM-as-a-judge.}
There is evidence of strong agreement of LLMs and human preferences in  grading outputs, with growing understanding of biases and mitigation \cite{Zheng2023LLMasJudge}.
\tts leverages LLM-as-a-judge constrained to \emph{pairwise} 
decisions at the chunk level. 
No prior RE work 
focuses on \emph{interview-to-story} alignment task with faithfulness/coverage.

\section{\tts: Interview-to-Story Alignment Task}
\label{sec:task}

We define the \textbf{interview-to-story alignment (\tts)} task as follows: given a transcript of an elicitation interview and a set of user stories derived from it, the goal is to quantify the extent to which the stories accurately reflect the information expressed by the participants during the interview. 

Formally, let $T$ be a transcript segmented into $n$ text chunks $C = \{c_1, c_2, \ldots, c_n\}$, and let $S = \{s_1, s_2, \ldots, s_m\}$ be a set of $m$ user stories. The goal is to determine, for each pair $(c_i, s_j)$, whether the chunk $c_i$ \textit{supports} the user story $s_j$, i.e., whether the requirement expressed in $s_j$ can be justified by information contained in $c_i$. 
We define two metrics:
\par\noindent
\textbf{Requirements Faithfulness}: the proportion of user stories supported by at least one chunk:
    \begin{equation}
    \text{Faithfulness} =
    \frac{\bigl|\{\, s_j \in S : \exists\, c_i \in C,\; y(c_i,s_j)=1 \,\}\bigr|}
         {|S|},
    \end{equation}
    where $y(c_i,s_j) \in \{0,1\}$ is the \textit{alignment label}, with 1 if the requirement in $s_j$ is justified by information in $c_i$, 0 otherwise. 
\par\noindent    
\textbf{Interview Coverage}: the proportion of chunks in the transcript that are covered by at least one story:
    \begin{equation}
    \text{Coverage} =
    \frac{\bigl|\{\, c_i \in C : \exists\, s_j \in S,\; y(c_i,s_j)=1 \,\}\bigr|}
         {|C|}.
    \end{equation}
\par\noindent

Coverage is a relative metric and may not reach 1.0, since not all transcript chunks necessarily express stakeholder needs. 
This should not be confused with requirements completeness, which is important but not covered by our heuristics.
%

{
Since interviews vary widely in length and structure, these metrics are intended for \emph{within-interview comparisons} (rather than across-interview metrics), such as comparing alternative story sets derived from the same transcript.} 

\section{Our Automated Approach to \tts}
\label{sec:method}
{We define our approach to solve the \tts task as a two-stage pipeline}: 
(1) segmenting the transcript into chunks, and 
(2) aligning the chunks with user stories via a matcher. 
An optional blocking stage can be inserted between these steps to reduce computational cost without altering the task definition. {The requirements faithfulness and interview coverage metrics are then derived from the alignment output.}

Formally, let $C = \{c_1, \ldots, c_n\}$ be the set of transcript chunks and $S = \{s_1, \ldots, s_m\}$ the set of user stories. 
We consider the set of candidate pairs
\[
P = C \times S = \{(c_i, s_j) \mid c_i \in C, s_j \in S \}.
\]

A matcher $X$ is a function
\[
X : C \times S \to \{0,1\},
\]
where $X(c_i, s_j) = 1$ indicates that chunk $c_i$ \emph{supports} story $s_j$. 
By default, $X$ is applied to the entire Cartesian product $P$, i.e., all possible pairs. 

\noindent The approach admits two operating modes:
\begin{enumerate}
    \item \textbf{Direct matching}: evaluate $X$ on all pairs $P = C \times S$.
    \item\textbf{Blocked matching}: apply an embedding-based blocking operator $B_K$ that restricts $P$ to a smaller candidate set $P'$, then evaluate $X$ only on $P'$.
\end{enumerate}

We now detail the three core components: 
\textit{chunking} ($C$), 
\textit{pairwise matching} ($X$), 
and \textit{embedding-based blocking} ($B_K$).

\subsection{Chunking ($C$)}
We segment transcripts based on speaker turns (as in Fig.~\ref{fig:ex}), following prior work on identifying requirements information in interviews~\cite{spijkman2023summarization}. While one could in principle treat each turn as a separate chunk, elicitation interviews are generally a question-and-answer sequence~\cite{zaremba2021towards}, which provides contextual information (spanning across multiple turns) that helps understand whether a user story is supported.

{Following the approach by previous RE research~\cite{spijkman2023summarization}, we adopt the strategy of grouping three consecutive speaker turns into a unit}, using a stride of one so that every possible three-turn window is considered and full interview coverage is ensured. Given turns $(t_1, t_2, \ldots, t_k)$, this results in overlapping chunks 
$(t_1,t_2,t_3), (t_2,t_3,t_4), \ldots, (t_{k-2},t_{k-1},t_k)$. 
%

\subsection{Pairwise Matching ($X$)}
\label{subsec:pairwise}
Given the chunk set $C$ and the user stories $S$, we form all possible pairs $(c_i, s_j)$. For each pair, a model $X$ assigns a binary label indicating whether the chunk supports the story. 

The matcher $X$ can be instantiated as:
\begin{enumerate}
    \item \textbf{LLM-based judge}, a large language model prompted with few shot examples of aligned and non-aligned pairs (see a prompt excerpt in Fig.~\ref{prompt1} and the full prompt in the online appendix: \url{https://zenodo.org/records/20596692});
    \item \textbf{Cross-encoder}, a reranker model that computes the similarity between $c_i$ and $s_j$ by jointly encoding $(c_i, s_j)$ and by returning a support score; or
    \item \textbf{Bi-encoder}, which computes embeddings for $c_i$ and $s_j$ independently and calculates their cosine similarity.
\end{enumerate}

In all cases, $X$ answers the question: \textit{“Does this interview chunk justify this user story?”}

\begin{tcolorbox}[colback=gray!5, colframe=gray!30, title=System Prompt, fonttitle=\bfseries, coltitle=black, boxrule=0.5pt, breakable]
You are a senior requirements analyst. \\
Your task: decide whether the \textbf{Chunk Text} gives \emph{enough evidence} to \emph{justify} the \textbf{User Story}. \\[4pt]

\textbf{Decision rules} \\
-- Output \texttt{1} if a knowledgeable reader could infer the User Story from the Chunk Text alone. \\
-- Otherwise output \texttt{0}. \\[4pt]

\textbf{Note:} Each Chunk Text is an excerpt from an informal interview transcript. 
Expect colloquial language, filler words (``uh'', ``you know''), and broken grammar; ignore these and focus on meaning. \\[4pt]

\textbf{Additional guidance} \\
-- User stories follow the format: \texttt{"As a <type of user>, I want to <goal> so that <reason>."} \\
-- Pay special attention to the \textbf{goal} in the User Story. \\
-- Verify that the \textbf{type of user} (e.g., ``employee'') is consistent with the Chunk Text. \\[4pt]

\textbf{Example (1 of 4)} \\[2pt]
\texttt{User Story: As a match official, I want to report on match events live, so that I do not register them twice.} \\
\texttt{Chunk Text: <excerpt about referee recording events in real time>} \\
Answer: \texttt{1} \\[2pt]
\texttt{... (three more few-shot examples omitted for brevity) ...} \\[6pt]

Now answer for the next pair. \\
Return \textbf{only one character}, either \texttt{1} or \texttt{0}, followed by nothing else.
\end{tcolorbox}
\noindent\begin{minipage}{\columnwidth}
\captionof{figure}{System prompt for the LLM-based judge matcher.}\label{prompt1}
\end{minipage}


\subsection{Embedding-Based Blocking ($B_K$)}
To reduce computational cost, we introduce an embedding-based blocking step. Instead of evaluating all pairs, we first compute embeddings for all chunks and stories and measure their similarity. For each story $s_j$, we keep only the Top-$K$ most similar chunks:
$
B_K(S,C) = \bigcup_{j=1}^{m} \{(c_i, s_j) : c_i \in \text{TopK}(C, s_j)\}
$.
The matcher $X$ is then applied only to these reduced candidate sets $P' = B_K(S,C)$. 
All other pairs are labeled as $X(c_i, s_j)=0$.
This blocking scheme preserves alignment quality while drastically reducing computation by a factor 
$\frac{|C|}{K}$.
For instance, with $|C| = 100$ and $K=25$, blocking cuts the number of  model calls by $4\times$.

\begin{table*}[!h]
\centering
\small
\caption{Statistics of the 17 datasets. 
Private datasets consist of student projects (each with two interviews). }
\label{tab:datasets_full}
\begin{tabular}{lrrrl}
\toprule
\textbf{Dataset} & \textbf{\#Stories} & \textbf{\#Chunks} & \textbf{$|C\times S|$ tokens} & \textbf{Domain} \\
\midrule
Student Project 1 & 53 & 386 & 3.76M & Predictive Dental Appointment Scheduler \\
Student Project 2 & 44 & 408 & 3.26M  & Centralized communication system for supermarkets \\
Student Project 3 & 59 & 546 & 3.87M & Information system for modernization of bakery \\
Student Project 4 & 80 & 169 & 2.61M & Learning support information system \\
Student Project 5 & 50 & 564 & 3.78M & Bike business improvement to kickstart efficiency \\
Student Project 6 & 62 & 380 & 4.73M & Workflow management system for solar panel company \\
Student Project 7 & 82 & 450 & 5.84M & Movie and series availability system \\
Student Project 8 & 46 & 277 & 2.52M & User-friendly app for devs to upload apps on Steam \\
Student Project 9 & 57 & 538 & 4.48M & Unified digital workspace for UX agency \\
Student Project 10 & 55 & 201 & 2.57M & Information system for automated train operation \\
Student Project 11 & 71 & 406 & 4.92M & Driving schools comparator \\
Student Project 12 & 52 & 714 & 4.36M & Cinema organization \\
Student Project 13 & 121 & 386 & 9.29M & Football league operations department \\
Student Project 14 & 64 & 440 & 4.49M & Scheduling tool for pizzeria \\
Student Project 15 & 39 & 429 & 2.39M & Venue management system events coordination \\
\midrule
Public Interview  & 37 & 132  & 0.88M & International football association portal \\
System Description  & 115 &37 & 0.19M & Travel agency management system \\
\bottomrule
\end{tabular}
\end{table*}

\section{Experimental Setup}
\label{sec:exp_setup}

 \noindent{\textbf{Datasets.}}
We evaluate our approach on 17 datasets (Table~\ref{tab:datasets_full}). 
The primary source consists of 15 software projects from a 
requirements engineering course. In each project, students conducted elicitation interviews while role-playing stakeholders and analysts to gather requirements for a target software system, following the pedagogical approach proposed by Ferrari \textit{et al.}~\cite{ferrari2020sapeer}.
For each project, we collected (i) transcripts of two elicitation interviews, concatenated into a single document, and (ii) the corresponding set of user stories. All data are in English and each project targets a distinct application domain. Students could refine requirements outside the elicitation sessions. Due to privacy constraints, these datasets cannot be released publicly and only local models can be used on them.

In addition, we include two public datasets:
\begin{enumerate}
    \item \textit{Public Interview Dataset}: a public interview transcript~\cite{spijkman2023summarization}, combined with a set of user stories~\cite{dalpiazextraction} for the same system, 
    forming a transcript--stories pair comparable to our private projects.

    \item \textit{System Description Dataset}: a textual system description with corresponding user stories
~\cite{santosuser}. 
\ul{Note}: While this dataset does not originate from interviews, we want to assess whether our method would also work for non-conversational data. Here, we cannot build chunks based on speaker turns, but instead split at newlines.
\end{enumerate}

To the best of our knowledge, no public dataset explicitly annotates which segments of a natural language source (interview transcripts or descriptions) support individual user stories. Existing work either evaluates user stories in isolation~\cite{yamani2025leveraging,quattrocchi2025can}, or compares generated stories against a reference set, without grounding them in the originating text~\cite{sharma2025evaluating,ronanki2022chatgpt}. 
Korn \textit{et al.}~
\cite{korn2025llmrei} mark whether high-level requirements are present in a transcript, but do not identify their exact location. In contrast, our annotated datasets explicitly target \emph{source grounding} by providing manually created alignments between user stories and corresponding chunks of the source text.

\vspace{.4ex}\noindent{\textbf{Annotation Protocol.}}
\label{subsec:annotation_protocol}
We manually annotated {matching pairs for} two private student projects (from different domains) and the public datasets (Table~\ref{tab:datasets_test}).
Exhaustively annotating all chunk-story pairs is infeasible due to the $|C|\times|S|$ complexity, so we adopted a similarity-driven protocol:
\begin{enumerate}
    \item For each user story, we retrieved source chunks ranked by decreasing embedding similarity using all-mpnet-base-v2;
    \item We selected the top-5 non-overlapping chunks (given stride-based chunking);
    \item If fewer than two positive matchings were found among the top-5, we extended the ranking until at least two positives were identified;
    \item When we knew from prior reading that a specific chunk was the most suitable support, we included it even if it was not in the top candidates.
\end{enumerate}

This process led to a {balanced yet feasible} set of annotated examples, {which are used exclusively to evaluate the pairwise chunk--story matching step.}


\begin{table}[!h]
\centering
\small
\caption{Statistics of the annotated test datasets used for evaluation. Each test set is a subset of the full dataset, selected and balanced via our annotation protocol.}
\label{tab:datasets_test}
\begin{tabular}{lrrr}
\toprule
\textbf{Dataset} & \textbf{\#Total} & \textbf{\#Positive} & \textbf{\#Negative} \\
\textbf{} & \textbf{Pairs} & \textbf{pairs} & \textbf{pairs} \\
\midrule
Student Proj. 1 (S1) & 287 & 109 & 178 \\
Student Proj. 2 (S2) & 236 & 88 & 148 \\
Public Int. (PI)& 180 & 74 & 106 \\
System Desc. (SD) & 580 & 302 & 278 \\
\bottomrule
\end{tabular}
\end{table}

We assessed inter-annotator reliability on a sample (50 chunk-story couples taken from two projects) using Fleiss’ $\kappa$ with three annotators. The overall agreement is $\kappa{=}0.470$ (\emph{moderate} agreement), indicating that agreement between humans is also challenging and opens up future directions (see Sect.~\ref{sec:discussion}). 
Additional details are in the online appendix.

\vspace{.4ex}\noindent{\textbf{Metrics}.}
We use the following criteria:
\begin{enumerate}
    \item  \textit{Macro F1-score}: evaluates the alignment quality of the matcher $X$ on the annotated datasets.
    \item \textit{Faithfulness and Coverage metrics}: assessed in different scenarios to demonstrate the usefulness of our approach.
    \item \textit{Computational cost}: measured as the total number of tokens processed (see $|C\times S|$ tokens in Table~\ref{tab:datasets_full}), allowing us to compare blocking and non-blocking approaches.
\end{enumerate}

\vspace{.4ex}
\noindent \textbf{Implementation.}
Our Python code is available at \url{https://zenodo.org/records/20596692}, also acts as online appendix, as it allows reproducing the experiments with the two public datasets. 

\section{Results}
\label{sec:results}
{
We first evaluate the quality of matchers for the pairwise matching task against human-annotated pairs (Sect.~\ref{subsec:results_align}). Then, we show that the proposed faithfulness and coverage metrics, computed over the full set of projects with the best matcher, effectively compare different sets of user stories, including both human- and LLM-generated ones (Sect.~\ref{subsec:correctness_completeness}). Next, we assess the efficiency of the embedding-based blocking strategy for the matching task (Sect.~\ref{subsec:results_block}). Finally, we compare the chunking approach versus giving the entire interview transcript to the matcher (Sect.~\ref{subsec:ablation}).}

\subsection{Pairwise Alignment Quality}
\label{subsec:results_align}
We evaluate the quality of the pairwise matcher $X$ (Sect.~\ref{subsec:pairwise}) on the manually annotated datasets, measuring macro F1-score across positive and negative pairs. 
\begin{table*}[b]
\centering
\small
\caption{Macro F1-scores of different matcher instantiations $X$ on the four annotated datasets. Results for threshold-based models are reported with best threshold and marked with [t]. Best score is in bold and second best is highlighted in grey.}
\label{tab:alignment_quality}
\begin{tabular}{lccccc}
\toprule
\textbf{Matcher ($X$)} & \textbf{Student 1} & \textbf{Student 2} & \textbf{Public Interview} & \textbf{System Description} & \textbf{Average} \\
\midrule
Bi-encoder [t] (Qwen3-0.6B) & 0.424 & 0.593 & 0.649 & 0.718 & 0.596 \\
Bi-encoder [t] (Qwen3-4B) & 0.362 & 0.576 & 0.741 & 0.741 & 0.605 \\
Bi-encoder [t] (Qwen3-8B) & 0.557 & 0.635 & 0.771 & 0.707 & 0.668\\
Cross-encoder [t] (Qwen3-Reranker-0.6B) & 0.647 & 0.679 & 0.748 & 0.841 & 0.729 \\
Cross-encoder [t] (Qwen3-Reranker-4B) & 0.765 & 0.728 & 0.797 & \textbf{0.914} & 0.801 \\
Cross-encoder [t] (Qwen3-Reranker-8B) & 0.776 & 0.737 & \second{0.829} & 0.895 & 0.809 \\
LLM (Qwen3-0.6B) & 0.310 & 0.365 & 0.291 & 0.507  & 0.406 \\
LLM (Qwen3-1.7B) & 0.326 & 0.597 & 0.483 & 0.693  & 0.525 \\
LLM (Qwen3-4B) & 0.774 & 0.676 & 0.732 & 0.856 & 0.760\\
LLM (Qwen3-8B) & 0.774 & 0.739 & 0.754 & 0.902 & 0.792 \\
LLM (Qwen3-14B) & 0.739 & 0.792 & 0.814 & 0.888 & 0.808 \\
LLM (Qwen3-32B) & \second{0.803} & \second{0.831} & 0.815 & \textbf{0.914} & \second{0.841} \\
LLM (Llama3.2-1B) & 0.455 & 0.418 & 0.493 & 0.324  & 0.423 \\
LLM (Llama3.2-3B) & 0.533 & 0.671 & 0.601 & 0.752  & 0.639 \\
LLM (Llama3.1-8B) & 0.578 & 0.696 & 0.703 & 0.863 & 0.711 \\
LLM (Llama3.3-70B) & \textbf{0.818} & \textbf{0.840} & \textbf{0.876} & \second{0.902} & \textbf{0.859} \\

\bottomrule
\end{tabular}
\end{table*}
Table~\ref{tab:alignment_quality} reports the results for the three matcher families described in Sect.~\ref{subsec:pairwise}: {LLM judges}, {cross-encoders}, and {bi-encoders}. All experiments are run with LLMs' temperature set to 0.

LLM judges emerge as the best-performing matchers. Qwen3-32B and Llama3.3-70B achieve macro-F1 scores above 0.80 across all datasets, with top average scores of 0.841 and 0.859, respectively. Cross-encoders deliver competitive results and are particularly effective at smaller model sizes; however, they output continuous scores and therefore require calibration to determine the optimal threshold that discriminates positives from negatives. For threshold-based approaches (bi-encoder and cross-encoder), Table~\ref{tab:alignment_quality} shows results for the best threshold per dataset, and should be interpreted as optimistic upper bounds. 
Finally, bi-encoder matchers perform worse overall, with low macro-F1 scores. 


\subsection{Stories' Faithfulness and Interview Coverage}
\label{subsec:correctness_completeness}

\begin{figure}[!h]
    \centering
    \includegraphics[width=0.95\columnwidth]{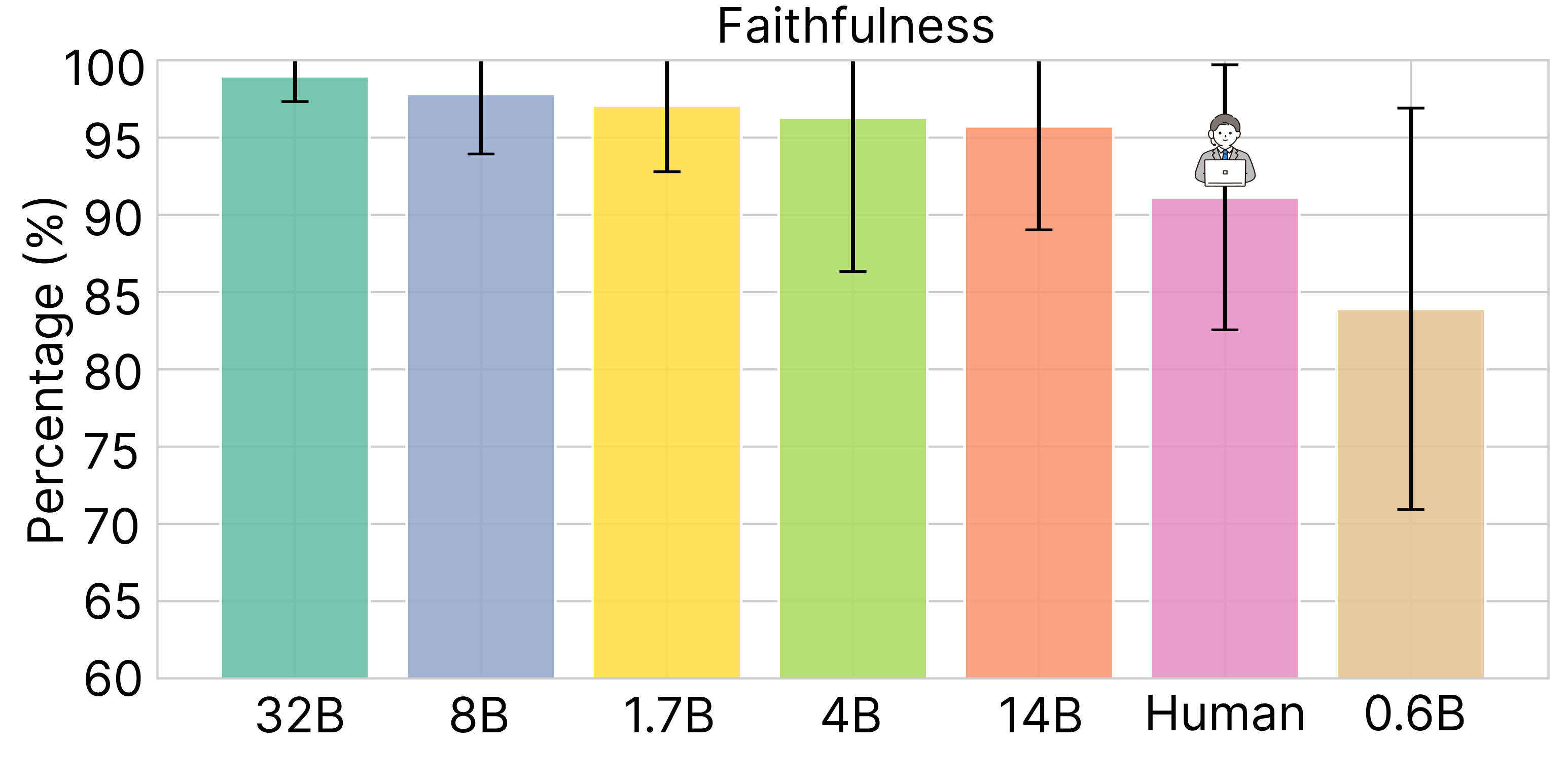}
    \caption{Average \textit{faithfulness} of user stories generated by Qwen models of increasing size.  Error bars denote $\pm$1 standard deviation across datasets (n=15).}
    \label{fig:correctness_plot}
\end{figure}

\begin{figure}[!h]
    \centering
    \includegraphics[width=0.95\columnwidth]{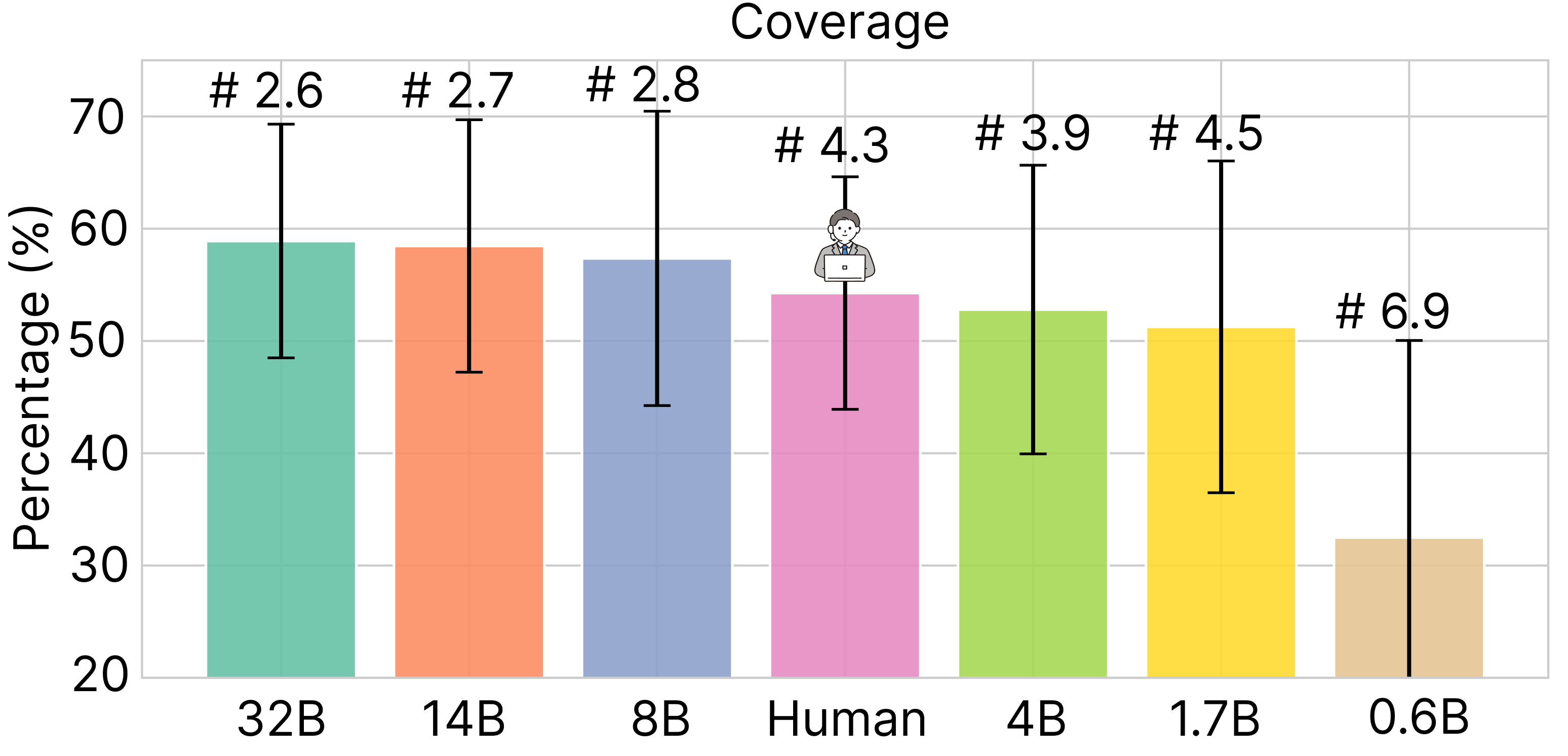}
    \caption{Average \textit{coverage} and {model ranking (\#)} of user stories generated by Qwen models of increasing size. Error bars denote $\pm$1 standard deviation across datasets (n=15).}
    \label{fig:completeness_plot}
\end{figure}

We now assess the behavior of our \textit{Faithfulness} and \textit{Coverage} metrics to validate them as indicators of source alignment. 
For each of the 15 private student projects, we let Qwen models of increasing size (0.6B to 32B) generate up to 50 user stories from the interview transcripts - the cap prevents excessively long generations and mitigates hallucinated ``never-ending'' outputs (see the online appendix for the generation prompt). We then apply {our} \tts\ {pipeline} in direct matching mode (no $B_K$), using Qwen3-32B as matcher $X$, and compute faithfulness and coverage.
Fig.~\ref{fig:correctness_plot} and Fig.~\ref{fig:completeness_plot} report the average scores across the 15 datasets, with standard deviation as error bars. We also report the metrics for the human-written stories produced by the students (\textit{Human}). {For coverage, we also report the average ranking (lower is better) across datasets.} 


{\textbf{Coverage increases with generator size} under a fixed judge and pipeline. This trend is expected: larger LLMs are known to generate more diverse and exhaustive outputs and should therefore cover a larger fraction of the interview content~\cite{kaplan2020scaling}. The monotonic increase across model sizes suggests that the metric is sensitive to this external signal, a pattern that is also reflected by the average coverage ranking across datasets reported in Figure~\ref{fig:completeness_plot}.}
LLM-generated story sets also achieve higher coverage than the human ones, as they tend to include more outputs, which humans may have overlooked. 

{
In contrast, \textbf{faithfulness is consistently high} across all but the smallest generator (Qwen3-0.6B). This asymmetry mirrors real-world software engineering practice: elicited requirements are often individually valid, but achieving complete requirements (here, approximated by coverage) is a recurring challenge due to the implicit and \textit{tacit} nature of requirements~\cite{sutcliffe2013requirements,ferrari2016ambiguity}. Interestingly, human-written stories obtain lower faithfulness than most LLM-generated sets. This is expected in our setting, as students were allowed to rely on sources beyond the interview transcripts, leading to stories not grounded in the interview data.}




\subsection{Blocking Efficiency}
\label{subsec:results_block}
We evaluate the extent to which the embedding-based blocking operator ($B_K$) reduces computational cost. We aim to obtain the same pairwise matching performance that would be achieved by applying a matcher to the full $|C|\times|S|$ Cartesian product, while processing fewer tokens.
Blocking must retain the pairs 
classified as positive by the matcher on the full cartesian product; if all positive pairs are preserved, downstream faithfulness and coverage scores remain unchanged.
We use Qwen3-32B as the matcher to compute the set of positive pairs from the full Cartesian product and we compare off-the-shelf Qwen3-Embedding-0.6B and its fine-tuned version. To fine-tune the embedding model, we generate training data by labeling all chunk-story pairs with the matcher and balancing positives and negatives via random downsampling. Evaluation follows a leave-one-out protocol across the 15 private projects.

\begin{figure}[h]
\centering
\includegraphics[width=1\columnwidth]{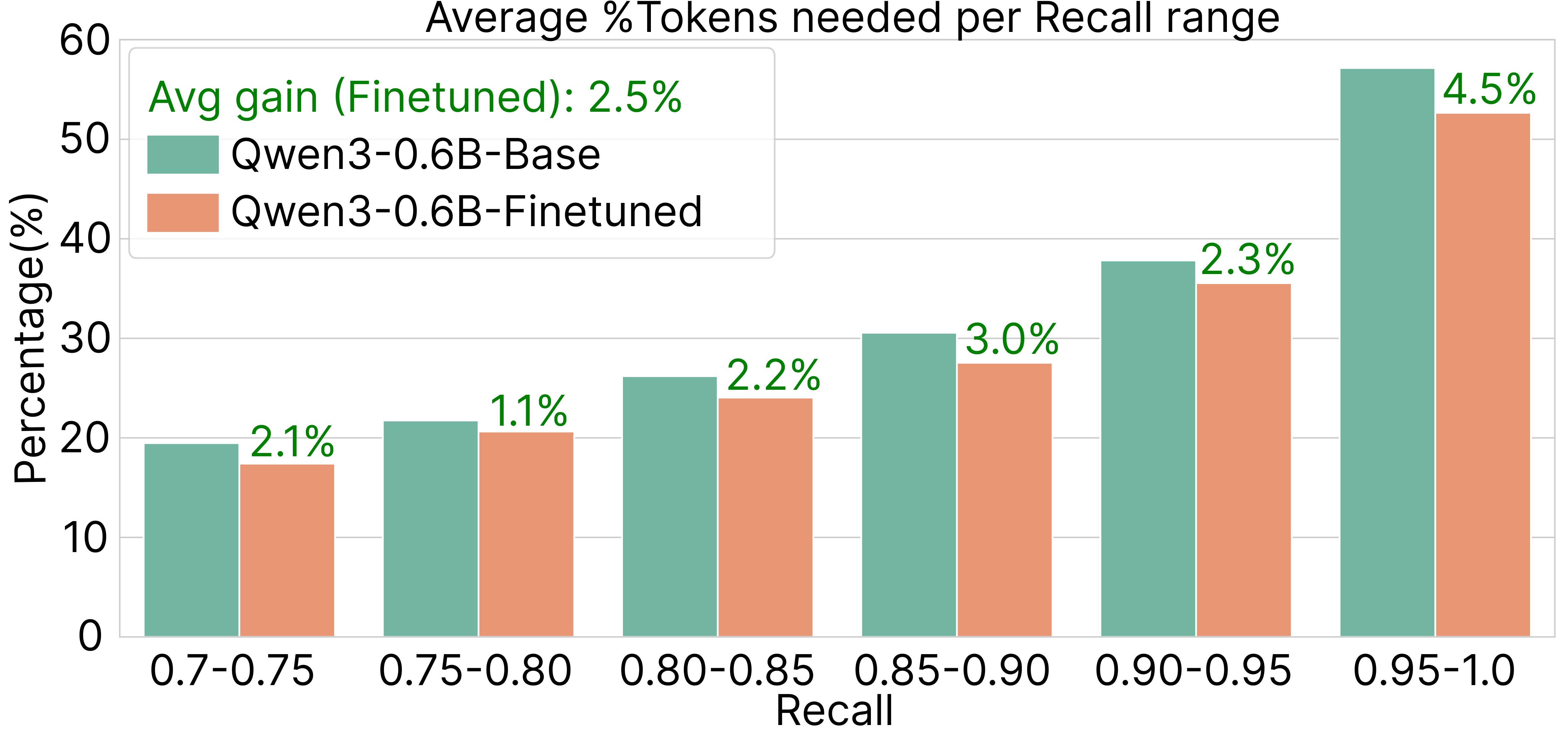}
\caption{Recall of positive pairs vs.~percentage of tokens retrieved by the blocking operator ($B_K$) across datasets. Lower values indicate higher efficiency.}
\label{fig:recall_vs_pct_tkns}
\end{figure}

Fig.~\ref{fig:recall_vs_pct_tkns} reports, for increasing recall levels, the fraction of tokens that must be processed after blocking.
For each story, we retrieve the top-$K$ most similar chunks and increase $K$ until the desired recall with respect to the reference positives is reached.
Because chunk lengths vary, we report token counts rather than $K$, expressed as a percentage of the Cartesian product. Blocking is efficient: on average, only one third of the tokens ($3\times$ reduction) are needed for recall 0.9–0.95, and about half ($2\times$ reduction) for recall 0.95–1. Fine-tuning further improves efficiency although not substantially. 



\subsection{Chunking vs. Feeding the Entire Interview}
\label{subsec:ablation}
Chunking is key to our alignment task, where the matching is between a fragment and a user story.  
To evaluate its impact, we compare this pairwise approach with an ablation setting in which the model receives the \emph{entire interview transcript}, segmented into numbered chunks (full-context). 
The model is asked, given a user story, to predict the indices of the supporting chunks. Transcripts are split into batches of 200 chunks to remain within context window limits. 

\begin{table}[h]
\centering
\small
\caption{Macro F1-scores comparing full-context evaluation with the pairwise matcher on the annotated datasets (S1, S2, PI, SD). Qwen3-32B is used as matcher $X$.}
\label{tab:chunking_ablation}
\begin{tabular}{lcccccc}
\toprule
\textbf{Dataset} & S1 & S2 & PI & SD & Avg \\
\midrule
\textbf{Full Context} & 0.471 & 0.540 & 0.623 & 0.322 & 0.489 \\
\textbf{Pairwise} & \textbf{0.803} & \textbf{0.831} & \textbf{0.815} & \textbf{0.914} & \textbf{0.841} \\
\bottomrule
\end{tabular}
\end{table}

As shown in Table~\ref{tab:chunking_ablation}, providing the model with full-context chunking degrades significantly the macro F1-score performance compared to the pairwise approach, thereby confirming the importance of chunking for the task at hand (given the limitations of existing language models). 

\section{Discussion and Roadmap}
\label{sec:discussion}
We introduced \tts, an evaluation framework that formalizes \emph{interview-to-story alignment} as a many-to-many matching between interview chunks and user stories. We build a pipeline that includes turn-aware chunking, a calibration-free pairwise judge, and an embedding-based blocking operator that preserves alignment recall while reducing token cost. 

{On four annotated sets, LLM judges attain strong alignment quality enabling the computation of two quality measures: {Requirements Faithfulness} and {Interview Coverage}. Applying these metrics to generated stories shows that they effectively assess both human- and LLM-written user stories.}

\subsection{Threats to Validity}
We discuss the main factors that may affect the results reported in this paper. Additional limitations are treated as opportunities and discussed in the roadmap.



Our \emph{interview coverage} is not equivalent to requirements completeness, which would be the ultimate metric to calculate in addition to faithfulness. 
A rigorous correlation study aligning transcript coverage with human expert-validated requirements would be valuable to establish to what extent our metric correlates with completeness.


Our metrics are not a universal measure of user story quality. 
\tts focuses on the \emph{source alignment} aspect and should be seen as a complement to existing quality frameworks such as QUS~\cite{Lucassen2016QUS}. 
While QUS criteria capture dimensions such as atomicity, estimability, uniqueness, and other syntactic, semantic and pragmatic criteria, our metric does not evaluate how well a story is written or whether it follows the user story template. 
{Currently, comparison with other metrics is difficult because there is no baseline for measuring how well a set of user stories reflects the content of stakeholder interviews.}

Our evaluation datasets are limited in scope. 
We rely primarily on student projects, which may not fully reflect industrial elicitation practices. 
Only two datasets are publicly available, both annotated in terms of matching by one of the authors and cross-checked with the other two on a subset, which may introduce annotation noise. 
Some experiments cannot be fully reproduced because certain transcripts cannot be released due to confidentiality constraints.
To keep manual annotation feasible, we used embeddings to limit the number of chunks to annotate; this may have led to an imperfect ground truth.

This paper does not aim at evaluating the effectiveness of prompt engineering, nor the suitability of certain prompts to specific models. We created simple prompts and we made them available. Nevertheless, we acknowledge that model performance could be affected by the selected prompts~\cite{ronanki2024requirements}.



\subsection{Roadmap}
The main contribution of this paper is that of formalizing a new NLP task for RE, besides classic ones~\cite{zhao2021natural} such as ambiguity detection, model generation, and trace link recovery.
The reported results are only a starting point and our research unveils numerous directions at the intersection of elicitation conversations and documented requirements.

\begin{description}
    \item[Aligning with other requirements and information types.] We only focused on user-oriented requirements, particularly through the notation of user stories. Although this kind of requirements is prevalent in agile development artifacts like backlog items~\cite{van2025locating}, there are other types of requirements, including non-functional ones, which could be traced back to their source. Furthermore, as shown by Spijkman \textit{et al.}~\cite{spijkman2021requirements}, requirements interviews may include additional information (like descriptions of as-is processes) that goes beyond requirements. Therefore, studying the matching of these additional requirements and information types would lead to a more comprehensive overview of alignment.
    \item[Requirements generation via LLMs.] We only touched the surface of a hot topic~\cite{quattrocchi2025can,korn2025llmrei}: the LLM-based generation of requirements. We did so in the context of evaluating the applicability of our metrics.
    The use of LLMs, and generative AI at large, for the goal of generating requirements is a very important direction, guided by questions concerning `How well do LLMs perform compared to humans?' To answer this type of questions, we need studies that focus on experimenting with a number of LLMs as well as prompting patterns, and that do not only take a quantitative standpoint, but that also account for qualitative differences (error profiles).
    Moreover, since requirements are only the inception of software development, it would be interesting to study how LLM-generated requirements affect downstream activities.
    \item[Revised Quality Frameworks.] While studying techniques for generating requirements from conversations, we expect revived interest in quality frameworks. We surmise that classical qualities of requirements, including adherence with a template, atomicity, or non-vagueness, can easily be achieved via language model via adequate prompting.
    Therefore, rather than focusing on these quality criteria, we would expect frameworks and automated tools that are able to focus on more advanced quality criteria that have to do with semantics, such as conflicts or difficult-to-estimate requirements. 

    \item[What ground truth?] 
    While advances in AI research are also supported by the establishment of ground-truth benchmarks, these are hardly existent when it comes to elicitation conversations and their requirements.
    First, very few requirements interviews are publicly available, and a subset of those have a corresponding specification.
    Second, there is no single set of requirements for a given conversation, and this makes alignment a complex challenge.
    As such, future research should explore in depth this issue, e.g., via user studies where multiple analysts and LLMs are required to generate requirements from the same interview.
    
    \item[Validation in industry.] We primarily analyzed conversations and requirements generated by groups of students in university projects.
    In addition to not being representative of the ability of a seasoned professional, all these interviews were conducted in a uniform setting: two interviews of roughly one hour each, focus on an information system, students instructed by the same lecturer, etc.
    We expect that validation in industry will reveal a high variety of interview types and of specification formats. This would call for approaches and evaluations that do not only consider the ability of generating \textit{one} type of requirements starting from \textit{one} kind of interview, but also their generality and adaptability.
\end{description}

\section*{Acknowledgement}
This work was partially supported by the AI4SWEng project, funded by the EU’s Horizon 
program (grant agreement No. 101189909), and by the 3IA Côte d’Azur Investments in the IA-cluster project managed by the  National Research Agency (ANR) with the reference number ANR-23-IACL-0001. We would like to thank the students who voluntarily provided their project data, thereby allowing our experiments.


\bibliographystyle{IEEEtran}
\bibliography{IEEEabrv,custom}

\end{document}